%% file: example_paper.tex
\documentclass{article}

\usepackage{microtype}
\usepackage{graphicx}
\usepackage{subfigure}
\usepackage{booktabs} % for professional tables
\usepackage{kotex}
\usepackage{amsmath,amssymb}
\usepackage[dvipsnames]{xcolor}
\usepackage{etoolbox}
\usepackage{breqn}
\usepackage{algorithm} 
\usepackage{algpseudocode}
\usepackage{setspace}
\usepackage{siunitx}
\usepackage{bm}
\usepackage{lipsum} % for dummy text
\usepackage{enumitem}

\usepackage{chngcntr}

\usepackage{caption}
\usepackage{comment}
\usepackage{makecell}

\usepackage{xfrac}
\usepackage{multirow}
\usepackage{mathtools}

\usepackage{hyperref}

\usepackage[accepted]{icml2021}

\icmltitlerunning{Submission and Formatting Instructions for ICML 2021}
\begin{document}

\twocolumn[
\icmltitle{Core-set Sampling for 
Efficient Neural Architecture Search}
% Conditional Focusing Learning in cGANs: A Closer Look at Density of Data Manifold \\ Enforcing High Density Samples for Conditional GANs\\ A Closer Look at Density of Data Manifold in Conditional GANs \\ CF-GAN: Conditional Focusing for Conditional GANs\\ SCF-GAN: Selective Conditional Focusing for Condtional GANs \\ Data Selective Focus for Improving Conditonal GANs Training \\ Selective Enforcing Learning for Conditional GANs: Enforcing High Density Samples to Generator \\ Improved training method for Conditional GANs: Enforcing High Density Samples to Generator\\Which Samples Conditional GANs Learns First?\\Instance concentration for Conditional GANs\\Concentration on Density of Distributions for Conditional GANs\\



\icmlsetsymbol{equal}{*}

\begin{icmlauthorlist}
\icmlauthor{Jae-hun Shim}{equal,Sogang}
\icmlauthor{Kyeongbo Kong}{equal,POSTECH}
\icmlauthor{Suk-Ju Kang}{Sogang}
\end{icmlauthorlist}

\icmlaffiliation{POSTECH}{Department of Electrical Engineering, POSTECH, Pohang, South Korea}
\icmlaffiliation{Sogang}{Department of Electronic Engineering, Sogang University, Seoul, South Korea}

\icmlcorrespondingauthor{Suk-Ju Kang}{sjkang@sogang.ac.kr}

\icmlkeywords{Machine Learning, ICML}

\vskip 0.3in
]



\printAffiliationsAndNotice{\icmlEqualContribution} % otherwise use the standard text.

\begin{abstract}
\input{0_abstract}
\end{abstract}

\section{Introduction}
\label{Introduction}
\input{1_introduction}

\section{Related Work}
\label{Related Work}
\input{5_relatedwork}

\section{Core-set Sampling for NAS}
\label{Proposed}
\input{3_method}

\section{Experiments}
\label{Experiments}
\input{4_experiments}

\section{Conclusion and Future Work}
\label{Conclusion}

\input{6_conclusion}

\section{Acknowledgements}
\label{Acknowledgements}
\input{9_Acknowledgements}

\bibliography{example_paper}
\bibliographystyle{icml2021}
\newpage

\end{document}

%% file: 0_abstract.tex
%With recent advances in Generative Adversarial Networks (GANs), it is possible to generate realistic images. However, it is still difficult to accurately model low density regions in the data manifold and hence, the generator of GAN often produces unrealistic samples. This paper focuses on analyzing and alleviating the instability problem that arises from sparse regions in the data manifold for conditional GANs (cGANs) which have class-wise controllability and superior quality. Our experiments show that the generator focuses on samples in low density regions with slower convergence, and this is caused by the conditional term of the discriminator having high confidence for samples of high density regions in the data manifold. 

Neural architecture search (NAS), an important branch of automatic machine learning, has become an effective approach to automate the design of deep learning models. However, the major issue in NAS is how to reduce the large search time imposed by the heavy computational burden. While most recent approaches focus on pruning redundant sets or developing new search methodologies, this paper attempts to formulate the problem based on the \textit{data curation manner}. Our key strategy is to search the architecture using summarized data distribution, i.e., core-set. Typically, many NAS algorithms separate searching and training stages, and the proposed core-set methodology is only used in search stage, thus their performance degradation can be minimized. In our experiments, we were able to save overall computational time from 30.8 hours to 3.5 hours, $8.8\times$ reduction, on a single RTX 3090 GPU without sacrificing accuracy.

%% file: 1_introduction.tex
Neural Architecture Search (NAS) has gained a wide attention in various deep learning tasks, such as image classification \citep{guo2020single, liu2018darts, wu2019fbnet}, object detection \citep{chen2019detnas, wang2020fcos, ghiasi2019fpn}, and segmentation \citep{liu2019auto,nekrasov2019fast}. The key objective of NAS is to replace human experts in designing the optimal architecture under constraints such as accuracy, memory, floating point operations (FLOPs), and latency. Early approaches \citep{tan2019mnasnet, liu2018progressive, zoph2018learning} failed in making the industry impact due to their large computational complexity. To reduce search time, most studies \citep{guo2020single, cai2018proxylessnas, wu2019fbnet} focus on reducing the candidate set, but too few candidate sets often do not guarantee the search of the best-performing architecture. 
Other studies \citep{pham2018efficient, liu2018darts} include utilizing transfer learning methodology. They search the architecture using a smaller proxy data such as CIFAR-10, and then, re-train the searched architecture using the full large-scale target task. Search time can be easily reduced using these methods, but there is no guarantee that the searched architecture is the optimal architecture for the target task due to the domain difference between search and training data. 

\begin{figure}[t]
\vskip -0.1in
\begin{center}
\centerline{\includegraphics[width=\columnwidth]{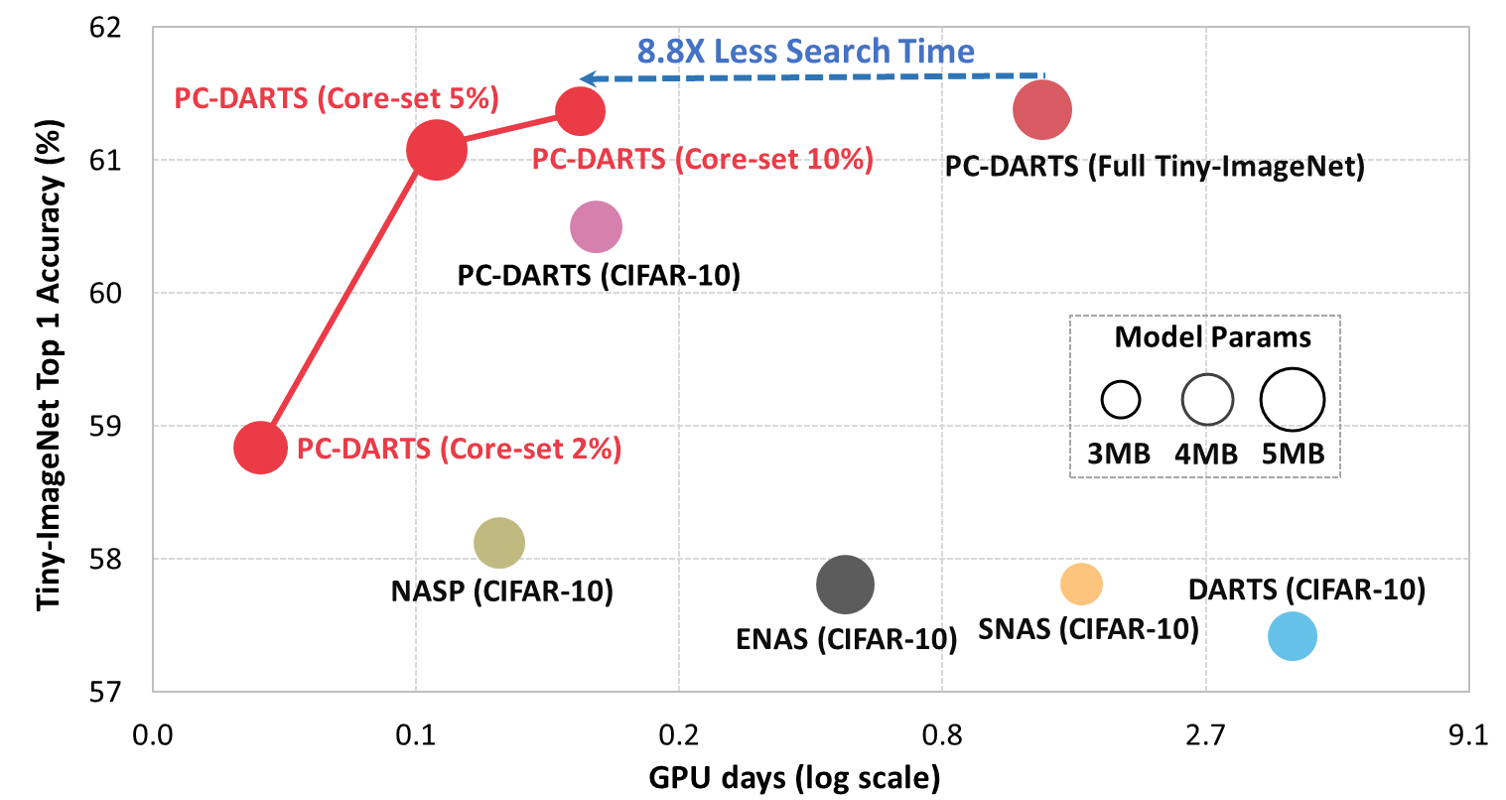}}
\vskip -0.15in
\caption{Comparison between models searched by our method and recent architectures on Tiny-ImageNet. Search time is reduced by 8.8 times with almost no loss in accuracy compared to model searched on full dataset. The size of the circles represents the number of parameters of each network. CIFAR-10 indicates that the model was first searched on CIFAR-10 and transferred to Tiny-ImageNet for training stage. Core-set indicates that the model was searched using the percentage of sampled data using our method.}
\label{ALL_NAS}
\end{center}
\vskip -0.2in
\end{figure}

This paper proposes a novel approach to use core-set selection \citep{agarwal2005geometric} in NAS. Our strategy is to search the architecture using core-set of samples that mimic the data distribution of the large target task. This strategy greatly reduces overall search time in exchange for a little reduction in search accuracy (Fig. \ref{ALL_NAS}). To select the core-set of the image manifold, images are first projected into an embedding space of perceptually-meaningful representations. A greedy k-center algorithm \citep{sener2018active} is then exploited to select samples with the $\textit{minimax facility location}$ formulation \citep{wolf2011facility}, using the Euclidean distance between the embedded data points. In the experiments, we evaluate a variety of image embeddings, observing that Inceptionv3 is appropriate for the task. Compared to the network searched on 10\% randomly sampled dataset, the model searched on same percentage of core-set sampled data achieved $2.1\%$ higher accuracy on Tiny-ImageNet dataset. Overall, we make the following contributions:

\vspace{-0.15in}
\begin{itemize}

\itemsep0em
\item We propose data curation methodology via the core-set selection to reduce the search time of NAS. To our best knowledge, it is the first work to adopt core-set selection for NAS.
\item We conduct experiments on the Tiny-ImageNet dataset and successfully show that our method can substantially increase accuracy compared to random data sampling for various ratios.
\item We exhibit the overall computational savings of core-set selection from 30.8 hours to 3.5 hours on a single RTX 3090 GPU by only searching on $10\%$ of the Tiny-ImageNet dataset.
\end{itemize}

%% file: 5_relatedwork.tex
%\textbf{Top-k, instance selection for GANs} Top-k: realistic sample에 더 집중하는 것과 비슷 (sprit 공유), top-k는 bad sample 학습에 제외, 우리는 good sample에 더 집중. 둘다 적용 가능하다, Instance selection: easy to classify sample과 high density sample은 거의 유사하다. instance selection은 data의 분포 관점에서 접근, 우리는 pattern관점에서 접근한 차이, 둘다 cGAN이고, 이놈도 결국 low density 제거하겟다는 거에서 topk와 비슷, 우리는 good sample에 집주이다!! 두개 같이 설명 가능할 듯

%\textbf{Mining GOLD Samples for Conditional GANs
%} 여기 gold example re-weghting과 자세히 비교하기.
%ac gan에 대해서 measure, 우리와 비슷하지만 학습 덜된 sample들에게 weight를 더주려는게 목표다.ㅣ 우리는 각 class별 easy sample들을 선별하여 conditional term으로 학습을 더 빠르게 만드는 것이 목표이기 때문에 차이가 있다. 저런 balance 문제도 없다. gold는 conditional과 unconditional에 같은 weight 적용, generatored sample에만 weight 적용
%gan이 학습이 잘된거를 판단하는 것이 아니라 학습 초반에 conditional term이 빠르게 학습되는 것을 이용하여 easy or high density sample 학습 가속화가 목표.

\subsection{Neural Architecture Search} 
Existing NAS methods can be split into following four categories in terms of the search strategy within the design space: the evolution-based method, the reinforcement-learning-based method, weight-sharing heuristics method and weight-sharing differentiable method.

First, evolution-based NAS \citep{liu2017hierarchical, xie2017genetic, real2017large} adopted evolutionary algorithms to build a family of architectures in the search phase, and architectures of similar properties with better performance are generated through mutation. Reinforcement-learning-based NAS, \citep{zoph2016neural, zhang2018unreasonable, baker2016designing, chen2019progressive} first proposed by \citep{zoph2016neural}, utilize a controller-based recurrent neural network to generate hyper-parameters using a  learnable policy. The generated hyper-parameters are then used to construct the network. However, both methods suffer from heavy cost in the search stage. Despite various efforts \citep{baker2016designing, chen2019progressive},  computational costs of these approaches are still unfit for general deployment.

The emergence of weight-sharing heuristic method \citep{pham2018efficient, brock2017smash, cai2018proxylessnas} has provided a solution for computational cost by constructing the search space into a super-network, from which various smaller networks can be sampled as a sub-network. With these methods, search time was reduced greatly as only one super-network training is required to search numerous sub-networks. However, sampled sub-networks with weights copied directly from the super-network do not guarantee the optimal performance as training procedure focuses on optimizing the entire super-network not individual sub-networks. To alleviate this issue, many methods adopt additional fine-tuning on the sampled sub-network.

Another way to conduct NAS is differentiable weight-sharing method, namely Differentiable Architecture Search (DARTS). In DARTS, continuous relaxation is used to construct the search space so that the super-network becomes differentiable to both the network weights and architecture parameters \citep{liu2018darts, liu2018progressive, xu2019pc}. Therefore, it is possible to use gradient update to jointly optimize model weights and architecture parameters, thereby producing the best performing architecture without an additional sub-network sampling procedure. Despite the efficiency of weight-sharing differentiable NAS, researches \citep{li2020random, yu2019evaluating} point out several weaknesses such as instability of architecture search due to the depth gap between the super-network and the sub-network, and computational overhead from computing gradients through a large and redundant search space. To tackle such issues, Progressive DARTS (P-DARTS) \citep{liu2018progressive} introduced a progressive search stage to bridge the depth gap, and Partially-Connected DARTS (PC-DARTS) \citep{xu2019pc} utilized partial channel connections in order to remove redundancy in the search space during the architecture search stage.

\subsection{Core-set Selection}
Core-set selection has been studied to find approximate solutions to the original NP-Hard problem \citep{agarwal2005geometric, clarkson2010coresets, pratap2018faster}. Core-sets have been applied to many machine learning problems including active learning for SVMs \citep{tsang2005core, tsang2007simpler}, unsupervised subset selection for hidden Markov models \citep{wei2013using}, and scalable Bayesian inference \citep{huggins2016coresets}. Recently, core-set selection was applied in the deep learning literature. \citep{sener2018active} and \citep{sinha2020small} utilized core-set selection for batch mode active learning and mimicking batch distribution in generative setting, respectively. In contrast, this is the first paper to apply core-set for NAS to reduce search time.

%% file: 3_method.tex
Our goal is to automatically select a subset that can minimize the architecture search performance degradation caused by the reduced number of data. To achieve this, we select the core-set of the data manifold, which can mimic the distribution of the target dataset. We first define an image embedding function $\psi(\cdot)$ and a distance metric $\Delta(\cdot,\cdot)$, and then, introduce a greedy $k$-center algorithm for NAS as the following sub-sections.

\subsection{Embedding Function and Distance Metric}  
Since the Euclidean distance in high-dimensional images is semantically meaningless \citep{girod1993s, eskicioglu1995image}, we need to project the images $\textbf{x}$ into the features $\psi(\textbf{x})$ of the low-dimensional space using the embedding function. For the search task, we use the feature space of a pre-trained image classifier that is a perceptually-aligned embedding function \citep{zhang2018unreasonable}. This paper uses the Inceptionv3 model pre-trained on the ImageNet dataset to extract the embeddings, and then, the Euclidean distance is used as the distance metric in the low-dimensional space.

\begin{algorithm}[t]
% \setstretch{1.35}
\AtBeginEnvironment{}{\setstretch{1}}
   \caption{Greedy $k$-center algorithm for NAS}
    \textbf{Input}: Training data $\textbf{x}$, total class $C$, selection ratio $r$, embedding function $\psi(\cdot)$, distance metric $\Delta(\cdot,\cdot)$
    \\
     \begin{algorithmic}
    \State \textbf{\#\,Core-set Selection in Data Curation Step}
    \State $Q$=\{\}
    \For {$c = 1, 2,..., C$}
        \State Fetch $\textbf{x}^c$ from $\textbf{x}$
        \State Initial set $\textbf{s}=\textbf{s}^0$ \quad\quad\quad\, \# Random initial set $\textbf{s}^0$
        \While{$|\textbf{s}|<=r\cdot|\textbf{x}^c|+|\textbf{s}^0|$}
        \State $u$ = arg$\max_{{x}^c_i\notin\textbf{s}}\min_{{x}^c_j\in \textbf{s}}\Delta(\psi(x^c_i)$, $\psi(x^c_j))$
        \State $\hat{x}^c = \psi^{-1}(u)$\quad \# Retrieve $\hat{x}^c$ from embeddeding
        \State $\textbf{s}=\textbf{s}\cup\{\hat{x}^c\}$
        \EndWhile
        \State $Q = Q\cup\{\textbf{s} \backslash \textbf{s}^0\}$
    \EndFor
    \\
    \State \textbf{\#\,Search Architecture}
    \State $\phi^0$ = Search($Q$)
    \\
    \State \textbf{\#\,Training Architecture}
    \State $\phi$ = Training($\phi^0$, $\textbf{x}$)
    \State \textbf{return} $\phi$
    \end{algorithmic}
   \label{algorithm2}
\end{algorithm}

\subsection{Core-set Selection}
In computational geometry, a core-set is a small set $Y$ that approximates the shape of an original set $X$ \citep{agarwal2005geometric}. Using the core-set for searching the architecture, we can generate approximated solutions with little computational complexity. To do so, we consider \textit{minimax facility location} formulation \citep{wolf2011facility} among the typical core-set selection problems:
\begin{equation}
\min_{Y:|Y|=k}\max_{x_i\in X}\min_{x_j\in Y}\Delta(x_i,x_j),
\label{eq:conditional_ratio}
\end{equation}
where $k$ is the desired size of $Y$, and $\Delta(\cdot,\cdot)$ is a distance metric. By this formulation, $k$ center points are chosen such that the largest distance between a data point and its nearest center is minimized. However, the exact solution of $k$-center problem is NP-Hard \citep{fowler1981optimal, megiddo1982complexity, gonzalez1985clustering}. Therefore, we instead use the greedy $k$-center algorithm which is also exploited in \citep{sener2018active,sinha2020small}. 

\subsection{Greedy $k$-center Algorithm for NAS}
The proposed method is described in Algorithm \ref{algorithm2}. First, for the entire training data $\textbf{x}$, embedded features are extracted from the pre-trained function $\psi(\cdot)$. Then, for \textit{each class}, we select the core-set using the greedy $k$-center algorithm that iteratively measures the distance between a pair of embedded features, and adds the sample $x_i$ that satisfies the minimax criterion until the selection ratio is reached. Using the selected core-set $Q$, differentiable NAS algorithms search the architectures. Among the various DARTS algorithms, we used the state-of-the-art PC-DARTS algorithm. The detailed searching method is described in Section \ref{Searching method}. Lastly, the searched architecture $\phi^0$ is trained on \textit{full training dataset} $\textbf{x}$. %The reason why core-set is used for search is described in Section \ref{Searching method}.

\begin{figure}[t]
\vskip -0.1in
\begin{center}
\centerline{\includegraphics[width=0.95\columnwidth]{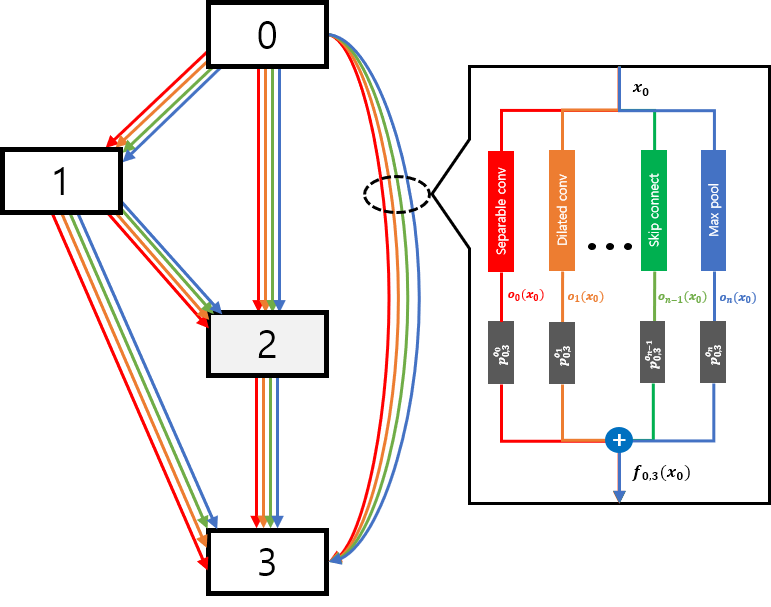}}
\vskip -0.15in
\caption{Illustration of DARTS cell search. Each box denote the nodes defined within a cell. Arrows denote the edges in the DAG structure. The diagram on the right hand side shows an edge connecting node 0 and 3. The output from node 0, $x_0$, propagates through the operators $\{o_0, o_1, \cdots, o_n\}$ and weighted sum of the outputs of each operators form the final output of the edge $f_{0,3}(x_0)$. The searched cell is stacked multiple times to form a network.}
\label{Darts_fig}
\end{center}
\vskip -0.2in
\end{figure}

%% file: 4_experiments.tex
In this section, we review search algorithms and analyze the impact of the proposed core-set selection for NAS.

\subsection{Search Algorithms}
\label{Searching method}
\subsubsection{DARTS} 

The goal of DARTS \citep{liu2018darts} is to search for a robust cell and apply it to a network of $L$ cells. Each cell is defined as a Directed Acyclic Graph (DAG) of $N$ nodes, and each node represents a layer in the network. We denote the operation space as $\mathcal{O}$ in which each element ${o(\cdot)}$ is a candidate operation (e.g. separable convolution, dilated convolution, skip-connect, and max-pool). For each edge ${(i,j)}$ consisting of two nodes $i$ and $j$, the input from node $i$ is passed onto node  $j$.  With this notion, the information is propagated through the weighted sum of architecture over each operation in $\mathcal{O}$, which can be formulated as
\begin{equation}
f_{i,j}(x_{i}) = \sum_{o\in O}p_{i,j}^{o}\cdot o(x_{i}), 
\end{equation}
where 
\begin{equation}
p_{i,j}^{o}=\frac{exp(\alpha_{i,j}^{o})}{\sum_{o'\in O}exp(\alpha_{i,j}^{o'})},
\end{equation}
$x_{i}$ is the output of the $i$-th node, and $\alpha_{i,j}^{o}$ is weight of candidate operator $o$ from node $i$ to $j$. The output of the entire cell is constructed by concatenating the outputs from all nodes (Fig. \ref{Darts_fig}).

In the search stage, the architecture parameters $\alpha_{i,j}^{o}$ are updated according to the following bi-level optimization problem as  
\begin{equation}
\min_{\alpha} \mathcal{L}_{val}{(w^{*}(\alpha),\alpha)},
\end{equation}
\begin{equation}
\text{s.t} \quad w^{*}(\alpha) = {\arg}\min_{w}\mathcal{L}_{train}{(w,\alpha)}, \nonumber
\end{equation}
where $\mathcal{L}_{train}$ and $\mathcal{L}_{val}$ represent training and validation loss respectively.
When the search ends, operation $o$ with the largest $\alpha_{i,j}^{o}$ is preserved on each edge.

\subsubsection{PC-DARTS} 
PC-DARTS \citep{xu2019pc} is a variant of DARTS with the use of partial channel connection for enhancement in memory efficiency. When we denote channel sampling mask as $S_{i,j}$, the new output function can be formulated as
\begin{multline}
f^{PC}_{i,j}(x_{i};S_{i,j})\\ = \sum_{o\in O}p_{i,j}^{o}\cdot o(S_{i,j}*x_{i})+(1-S_{i,j})*x_{i},
\end{multline}
where $S_{i,j}*x_{i}$ and $(1-S_{i,j})*x_{i}$ denote the selected and masked channels, respectively. The portion of the selected channel is set as a hyper-parameter $q$, thus selecting $1/q$ channels for each operation.

Another method introduced in PC-DARTS is the edge normalization, which is used to mitigate the undesired fluctuation due to the random sampling of channels across iterations. With edge normalization on each edge ${(i, j)}$, denoted as $\beta_{i,j}$, computation of $x_{j}$ becomes the following equation. 
\begin{equation}
x^{PC}_{j} = \sum_{i<j}\frac{exp(\beta_{i,j})}{\sum_{i'<j}exp(\beta_{i',j})}\cdot f_{i,j}(x_{i}).
\end{equation}

When the search is finished, the connectivity of edge ${(i,j)}$ is determined by both $\frac{exp(\beta_{i,j})}{\sum_{i'<j}exp(\beta_{i',j})}$ and $\frac{exp(\alpha_{i,j}^{o})}{\sum_{o'\in O}exp(\alpha_{i,j}^{o'})}$.

\begin{table}[t]
\centering
\vskip -0.05in
\caption{Comparison of embedding function for selecting core-set 10\%. Best results in bold.}
\vskip 0.05in
\label{embedding_result}
\scalebox{0.8}{
\begin{tabular}{c|ccccc}
\hline
\multirow{2}{*}{\begin{tabular}[c]{@{}c@{}}\textbf{Selection}\\ \textbf{Method}\end{tabular}} & \multirow{2}{*}{\textbf{Network}} & \multirow{2}{*}{\textbf{Pre-training}} & \multicolumn{2}{c}{\textbf{Test Acc}} & \multirow{2}{*}{\begin{tabular}[c]{@{}c@{}}\textbf{Params}\\ \textbf{(MB)}\end{tabular}} \\ \cline{4-5}
                                                                            &                            &                               & \textbf{Top 1}              & \textbf{Top 5}             &                                                                        \\ \hline
\multirow{4}{*}{\begin{tabular}[c]{@{}c@{}}Core-set\\ 10\%\end{tabular}}    & Inceptionv3                & ImageNet                      & \textbf{61.4}               & \textbf{79.4}              & 3.9                                                                    \\
                                                                            & ResNet-50                  & ImageNet                      & 60.5               & 79.3              & 4.1                                                                    \\
                                                                            & ResNet-50                  & Places365                     & 58.5               & 77.1              & 2.7                                                                    \\
                                                                            & ResNet-50                  & SwAV                          &   56.9                 &     76.1              & \textbf{2.4}                                                                       \\ \hline
\end{tabular}}
\vskip -0.15in
\end{table}

\begin{table*}[t]
\centering
\caption{Comparison with different Selection Ratio (SR) on Tiny-ImageNet with PC-DARTS. Selected samples are only exploited in search stage. The searched architecture is trained using full dataset. Best results in bold.}
\vskip 0.05in
\label{Selection_Ratio}
\scalebox{0.9}{
\begin{tabular}{c|ccccccc}
\Xhline{2\arrayrulewidth}
\multirow{2}{*}{\textbf{Selection Method}} & \multirow{2}{*}{\textbf{SR}} & \multicolumn{2}{c}{\textbf{Test Accuracy (\%)}} & \multirow{2}{*}{\textbf{Params (MB)}} & \multicolumn{3}{c}{\textbf{Search Cost (GPU-hours)}}           \\\cline{3-4}\cline{6-8}
                                  &                         & \textbf{Top 1}           & \textbf{Top 5}           &                              & \textbf{Core-set Selection} & \textbf{Architecture Search} & \textbf{Total Cost} \\ \hline
\multirow{3}{*}{Core-set}         & 2\%                      & 58.8            & 77.9            & 4.2                          & \textbf{0.1}                & \textbf{0.7}                 & $\textbf{0.8}$        \\
                                  & 5\%                      & 61.1            & 79.5            & 4.8                          & 0.2                & 1.6                 & 1.8        \\
                                  & 10\%                     & $\textbf{61.4}$            & 79.4            & 3.9                          & 0.4                & 3.1                 & 3.5        \\ \hline
\multirow{4}{*}{Random}           & 2\%                      & 57.6            & 76.8            & 4.3                          & -                  & 0.7                 & 0.7        \\
                                  & 5\%                      & 59.6            & 78.5            & 4.0                          & -                  & 1.6                 & 1.6        \\
                                  & 10\%                     & 59.3            & 78.9            & \textbf{3.3}                          & -                  & 3.1                 & 3.1        \\
                                  & 50\%                     & 61.2            & 79.5            & 4.2                          & -                  & 15.3                & 15.3       \\ \hline
Full Dataset                      & 100\%                    & $\textbf{61.4}$            & $\textbf{79.9}$            & 4.7                          & -                  & 30.8                & 30.8       \\ \Xhline{2\arrayrulewidth}
\end{tabular}}
\vskip -0.05in
\end{table*}

\begin{table*}[t]
\centering
\caption{Comparison with recent architectures on Tiny-ImageNet. Each architecture was first searched on full CIFAR-10 or core-set 10\% of Tiny-ImageNet, and then, trained full Tiny-ImageNet for training stage. Best results in bold and $^{\ddagger}$ is quoted from
\citep{yao2020efficient}.}
\vskip 0.05in
\label{recent_architectures}
\scalebox{0.9}{
\begin{tabular}{c|cccccc}
\Xhline{2\arrayrulewidth}
\multirow{2}{*}{\textbf{Search Dataset}}                                                    & \multirow{2}{*}{\textbf{Architecture}} & \multicolumn{2}{c}{\textbf{Test Accuracy (\%)}} & \multirow{2}{*}{\textbf{Params (MB)}} & \textbf{Search Cost} & \multirow{2}{*}{\textbf{Search Method}} \\ \cline{3-4}
                                                                                   &                               & \textbf{Top 1}              & \textbf{Top 5}             &                              &  \textbf{(GPU-days)}                                        & \multicolumn{1}{l}{}                               \\ \hline
\multirow{8}{*}{\begin{tabular}[c]{@{}c@{}}CIFAR-10\\ (Full dataset)\end{tabular}} & ResNet18$^{\ddagger}$ \citep{he2016deep}                     & 52.7               & 76.8              & 11.7                         & -                                        & manual                                             \\
                                                                                   & NASNet-A$^{\ddagger}$ \citep{zoph2018learning}                      & 59.0               & 77.9              & 4.8                          & 1800                                     & RL                                                 \\
                                                                                   & AmoebaNet-A$^{\ddagger}$ \citep{real2019regularized}                  & 57.2               & 77.6              & 4.2                          & 3150                                     & evolution                                          \\
                                                                                   & ENAS$^{\ddagger}$ \citep{pham2018efficient}                         & 57.8               & 77.3              & 4.6                          & 0.5                                      & RL                                                 \\
                                                                                   & DARTS$^{\ddagger}$ \citep{liu2018darts}                        & 57.4               & 76.8              & 3.9                          & 4                                        & gradient-based                                     \\
                                                                                   & SNAS$^{\ddagger}$ \citep{xie2018snas}                         & 57.8               & 76.9              & \textbf{3.3}                          & 1.5                                      & gradient-based                                     \\
                                                                                   & NASP$^{\ddagger}$ \citep{yao2020efficient}                          & 58.1               & 77.6              & 4.0                          & \textbf{0.1}                                      & gradient-based                                     \\
                                                                                   & PC-DARTS \citep{xu2019pc}                      & 60.5               & 79.0              & 4.1                          & \textbf{0.1}                                      & gradient-based                                     \\ \hline
\begin{tabular}[c]{@{}c@{}}Tiny-ImageNet\\ (Core-set 10\%)\end{tabular}            & PC-DARTS \citep{xu2019pc}                     & \textbf{61.4}               & \textbf{79.4}              & 3.9                          & \textbf{0.1}                                      & gradient-based                                     \\ \Xhline{2\arrayrulewidth}
\end{tabular}}
\end{table*}

\subsection{Dataset and Implementation Details}
%We perform experiments on CIFAR10 and ImageNet, two most popular datasets for evaluating neural architecture search. CIFAR10 (Krizhevsky & Hinton, 2009) consists of 60K images, all of which are of a spatial resolution of 32  32. These images are equally distributed over 10 classes, with 50K training and 10K testing images. ImageNet (Deng et al., 2009) contains 1;000 object categories, and 1:3M training images and 50K validation images, all of which are high-resolution and roughly equally distributed over all classes. Following the conventions (Zoph et al., 2018; Liu et al., 2019), we apply the mobile setting where the input image size is fixed to be 224  224 and the number of multi-add operations does not exceed 600M in the testing stage. Following DARTS (Liu et al., 2019) as well as conventional architecture search approaches, we use an individual stage for architecture search, and after the optimal architecture is obtained, we conduct another training process from scratch. In the search stage, the goal is to determine the best sets of hyper-parameters, namely

We conducted experiments on Tiny-ImageNet \citep{le2015tiny} to verify the effectiveness of our method. Tiny-ImageNet is a subset of ILSVRC2012 ImageNet dataset \citep{russakovsky2015imagenet} that consists of 200 classes with 64$\times$64 pixel resolution. It contains 100K training images, 10K test images, and 10K validation images, respectively. Searching on Tiny-ImageNet is a much more complicated and burdensome task compared to searching on CIFAR-10, which has 50K training images, and 10K testing images with 32$\times$32 pixel resolution. Following the conventional DARTS, we decoupled the NAS process into search and training stages. In the search stage, training set was used for updating the network weight, and validation set was used to update the architecture parameters {$\alpha_{i,j}^{o}$} and  {$\beta_{i,j}$}. 

For the search stage, we generally followed the ImageNet search stage settings provided by PC-DARTS \citep{xu2019pc} except for the minor adjustment in the network architecture. We kept the first convolution layer of the network to the CIFAR-10 network settings instead of the ImageNet settings. The original ImageNet search settings added additional convolution layers to downsample the large resolution of ImageNet data, which is unnecessary in our case as we are dealing with Tiny-ImageNet with 64$\times$64 pixel resolution. 8 cells each consisting of N = 6 nodes, were stacked to construct the network with the initial channel number of 16. The network was trained for 50 epochs and the architecture parameters were frozen for the first 35 epochs. The channel sampling rate $1/q$ was set to $1/2$. For updating the network weights, we used SGD with initial learning rate of 0.1 annealing down to zero using a cosine scheduler, a momentum of 0.9, and a weight decay of $3\times10^{-5}$. Adam optimizer with learning rate of $6\times10^{-3}$, a momentum $\beta$ = (0.5, 0999), and a weight decay of $10^{-3}$ was used for updating the architecture parameters. The search was conducted on a single RTX 3090 with batch size of 64.

For the training stage, we emulated the Tiny-ImageNet training settings used by \citep{yao2020efficient} for fair comparison.  The same SGD optimizer settings in the search stage was used to update the network weights, along with an auxiliary loss tower previously implemented in the original DARTS \citep{liu2018darts}. 14 cells were stacked with the initial channel number of 48 to construct the network and it was trained for 250 epochs using a batch size of 96.

\subsection{Results on Different Embedding Function}
To understand the dependency of the embedding function, we compare various model embeddings that are trained on different datasets: Inceptionv3 \citep{szegedy2016rethinking} trained on ImageNet, ResNet50 \citep{he2016deep} trained on ImageNet, Places365 \citep{zhou2017places}, and with SwAV unsupervised pre-training \citep{caron2020unsupervised}. All features were extracted after the global average pooling layer. We selected 10\% of the training data using greedy $k$-center algorithm for various embedded features. In Table \ref{embedding_result}, the experimental results show better search results for core-set selected from ImageNet pre-trained embeddings compared to those selected using Places365 or SwAV pre-trained embeddings. This means that embedding features that are well aligned with the training data are required for effective core-set selection. In addition, embedding with Inceptionv3 is more effective than those with ResNet-50. Therefore, we chose Inceptionv3 pre-trained on ImageNet as the embedding function for the rest of our experiments.

\subsection{Results on Different Selection Ratio}
Selection ratio is also important in core-set selection for NAS. To understand the impact of selection ratio, we conducted core-set selection with ratios of $2\%$, $5\%$, and $10\%$. We also conducted random selection with ratios of $2\%$, $5\%$, $10\%$, and $50\%$ for comparison. Note that selected samples are exploited only in the search stage. After the search stage, the architecture was trained using the full dataset. In Table \ref{Selection_Ratio}, as selected portions of the training dataset increased, test accuracy and search time also increased concurrently. For all sampling ratios, the proposed core-set selection outperformed (up to $2.07\%$) random selection. Interestingly, the searched architecture using the full dataset and 10\% core-set achieved similar accuracy. This means that search time can be reduced by $8.8\times$ without sacrificing performance. Note that time cost for core-set selection is minimal compared to the search time for all sampling ratios.

\subsection{Results on Tiny-ImageNet}
In Table \ref{recent_architectures}, we compared the performance of core-set selection to conventional NAS algorithms. Conventional algorithms first searched the architecture using a CIFAR-10, and re-trained the searched architecture using the Tiny-ImageNet because of the large search time. Compared to conventional NAS algorithms, the proposed core-set selection had the highest test accuracy with only 0.15 GPU-days. This result shows that core-set selection for the target dataset is more effective than transfer learning methodology in terms of accuracy and search cost.

%% file: 6_conclusion.tex
This paper proposed an efficient data selection method for NAS via core-set selection. The main idea was to perform architecture search using a core-set that emulates the distribution of the target dataset. The $k$-center greedy algorithm was used to select the core-set by calculating the Euclidean distance between embedding features from a meaningful representation embedding space. After evaluating on various image embeddings, we settled on Inceptionv3 pre-trained on ImageNet as the most appropriate model for our task. We performed thorough experiments on Tiny-ImageNet dataset to prove the efficiency of our method. Our method successfully searched out a model comparable in performance to the model searched on full Tiny-ImageNet dataset in only 0.15 GPU days.

Despite the efficiency of data curation, it is largely underexplored in the field of NAS. PC-DARTS \citep{xu2019pc} utilizes 10\% random sampling on ImageNet dataset, and Data-Adaptive NAS (DA-NAS) \citep{dai2020data} progressively adds data from easy class to hard class during the search stage. However, as of our knowledge, we are the first to utilize the subset selection methodology by analyzing the full distribution of data to conduct NAS. As we have shown the effectiveness of core-set selection on NAS, we hope to expand onto a more efficient data sampling method to enhance its performance through future work.

Recently, weight-sharing heuristics NAS method aggregated attention due to its high performance despite its efficiency. However, in this method, the super-network weight is directly inherited to the sub-networks. Therefore, decoupling search and train stages is not so trivial as in differentiable methods, requiring a different approach in efficient data selection. In the future work, we hope to explore a dynamic data curation, including curriculum learning, that can be applied to a larger variety of NAS techniques.

%Our work may only be a small step, but hopefully it will become a cornerstone for a new field in NAS.

%추후 뒤쪽 결론에 Limitation에 dataset이 확장의 문제 등에 대해서 명시적으로 언급하는게 좋겠음.. future work로 하겠다는 것도..

%and achieved the state-of-the-art performance (in terms of FID) on both datasets
% we hope to explore other discriminator for cGANs by applying our method and find the optimal solution of SFL.

%% file: 9_Acknowledgements.tex
This research was supported in part by Basic Science Research Program through the National Research Foundation of Korea (NRF) funded by the Ministry of Education (2020R1A6A3A01098940, 2021R1I1A1A01051225) and the NRF grant funded by the Korea government (MSIT) (2021R1A2C1004208).